\def\year{2019}\relax
\documentclass[letterpaper]{article} 
\usepackage{aaai19}  
\usepackage{times}  
\usepackage{helvet}  
\usepackage{courier}  
\usepackage{url}  
\usepackage{graphicx}  
\usepackage{verbatim}

\begin{document}
\title{Solving Service Robot Tasks: UT Austin Villa@Home 2019 Team Report}
\author{
Rishi Shah\textsuperscript{1}, Yuqian Jiang\textsuperscript{1}, Haresh Karnan\textsuperscript{2}, Gilberto Briscoe-Martinez\textsuperscript{3}, Dominick Mulder\textsuperscript{2}, \\
\bf{ \Large Ryan Gupta\textsuperscript{3}, Rachel Schlossman\textsuperscript{2}, Marika Murphy\textsuperscript{1}, Justin W. Hart\textsuperscript{1}, Luis Sentis\textsuperscript{3}, Peter Stone\textsuperscript{1}} \\
\textsuperscript{1}Department of Computer Science, University of Texas at Austin, Austin, USA\\
\textsuperscript{2}Department of Mechanical Engineering, University of Texas at Austin, Austin, USA\\
\textsuperscript{3}Department of Aerospace Engineering and Engineering Mechanics, University of Texas at Austin, Austin, USA\\
\{rishihahs, jiangyuqian, haresh.miriyala, gilbert.martinez675\}@utexas.edu,\\
\{dominickm611, ryangupta8\}@gmail.com, \{rachel.schlossman, marikamurphy\}@utexas.edu,\\
hart@cs.utexas.edu, lsentis@austin.utexas.edu, pstone@cs.utexas.edu
}
\maketitle

\begin{abstract}
RoboCup@Home is an international robotics competition based on domestic tasks requiring autonomous capabilities pertaining to a large variety of AI technologies. Research challenges are motivated by these tasks both at the level of individual technologies and the integration of subsystems into a fully functional, robustly autonomous system. We describe the progress made by the UT Austin Villa 2019 RoboCup@Home team which represents a significant step forward in AI-based HRI due to the breadth of tasks accomplished within a unified system. Presented are the competition tasks, component technologies they rely on, our initial approaches both to the components and their integration, and directions for future research.

\end{abstract}

\section{Introduction}

RoboCup@Home is an international robotics competition dedicated to advancing the state of the art in human-robot interaction for service robots.  The 2019 event in Sydney, Australia was the first using a new rulebook defining a variety of domestic tasks requiring autonomous capabilities pertaining to a large variety of AI technologies including visual object recognition, person recognition and tracking, navigation, manipulation, natural language understanding and generation, knowledge representation and reasoning, and planning.  Research challenges abound, both at the levels of these individual technologies, and especially at their integration into a fully functional, robustly autonomous system.

This paper describes the UT Austin Villa 2019 RoboCup@Home team, a collaborative effort between PIs and students in the Computer Science, Mechanical Engineering and Aerospace Engineering departments at the University of Texas at Austin.  While we were unable to demonstrate our capabilities fully at the competition itself due to hardware issues,\footnote{The robot we shipped to the competition did not clear customs in time, and the replacement we were loaned had some faulty hardware.} the progress made leading up to the competition represents a significant step forward in AI-based HRI due to the breadth of tasks accomplished within a unified system. The robot used in Robocup@Home Domestic Standard Platform League tasks is a Toyota Human Support Robot (HSR).  

Competing with the new rules for the first time, we aimed to take the simplest possible approach to each task, and as such, many challenges remain.  The main contribution of this paper is to describe the tasks we attempted in the competition, the component technologies they rely on, our initial approaches both to the components and their integration, and directions for future research.
In each case, we document, through videos and/or descriptions, our most successful trials, as a way of demonstrating both progress made to date and the room for improvement in future years.  We particularly emphasize the AI-HRI aspects of the challenges.

\section{The Standard Platform Robot}

\begin{figure}[t]
    \centering
    \includegraphics[width=0.6\columnwidth]{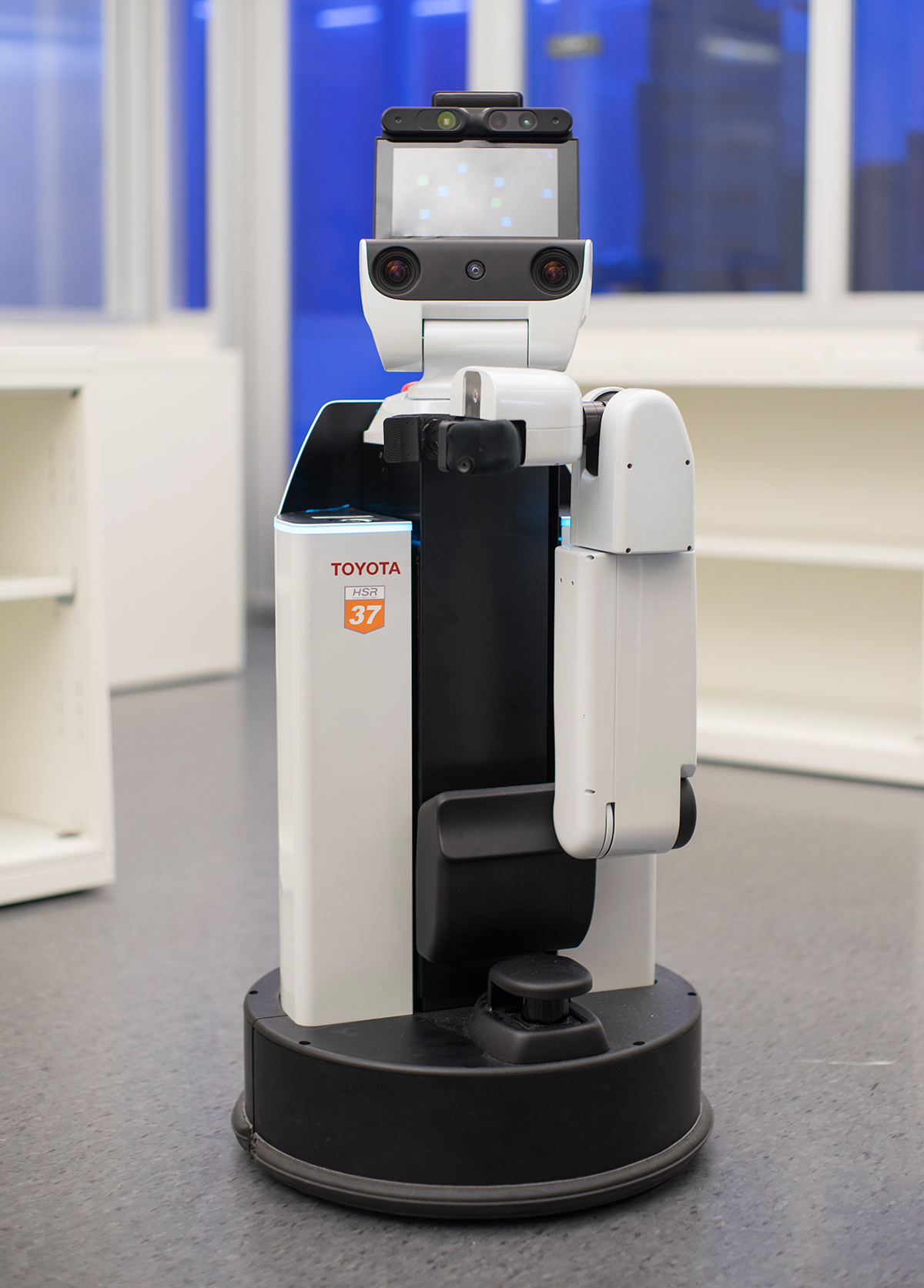}
    \caption{The HSR Robot.}
    \label{fig:hsr}
\end{figure}

RoboCup@Home includes three different subleagues, one of which requires designing the robot itself, and the other two of which use standard platforms such that all teams have exactly the same hardware.  UT Austin Villa participates in the Domestic Standard Platform League (DSPL), in which all teams use the Toyota HSR robot~\cite{Yamamoto2019}. Figure \ref{fig:hsr} shows the HSR, which has a holonomic omnidirectional base, a 4 DoF arm, a 1 DoF torso lift, a head RGBD sensor, and an RGB fisheye hand camera. For fast neural network computation, an external laptop with an Nvidia RTX 2070 GPU is mounted to the robot.
The competition thus amounts to a software competition:  the raw perception and actuation capabilities available to all the teams is identical.

\section{Task Descriptions}

This section roughly describes the tasks we attempted in the competition, and identifies the AI component technologies needed for each one.  For full task descriptions, we refer the reader to the RoboCup@Home rulebook ~\cite{rulebook2019}.

\begin{figure}[t]
    \centering
    \includegraphics[width=0.9\columnwidth]{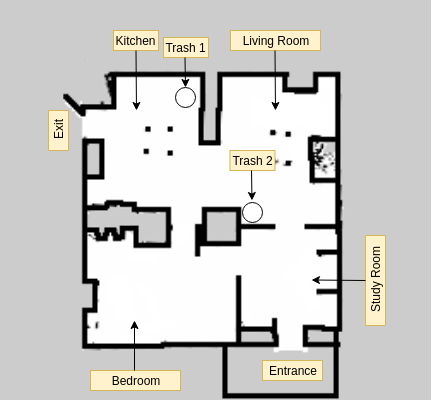}
    \caption{The 2019 competition arena mapped by a Toyota Human Support Robot.}
    \label{fig:arena}
\end{figure}

All of the tasks take place in a mock apartment (“the arena”) consisting of 4 rooms furnished as a living room, a kitchen, a study, and a bedroom. There are two doors to the exterior. An example arena is pictured in Figure~\ref{fig:arena}.  The robot is able to map the arena prior to performing any of the tasks.

\begin{figure}[t]
    \centering
    \includegraphics[width=0.9\columnwidth]{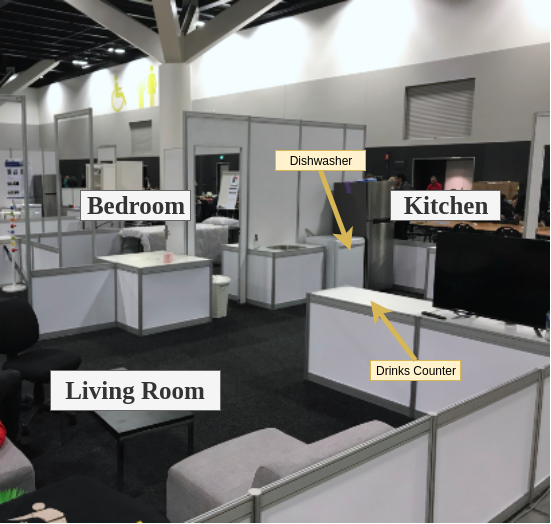}
    \caption{The mock apartment setup as seen from the Living Room.}
    \label{fig:arena_photo}
\end{figure}

\begin{figure}[t]
    \centering
    \includegraphics[width=0.9\columnwidth]{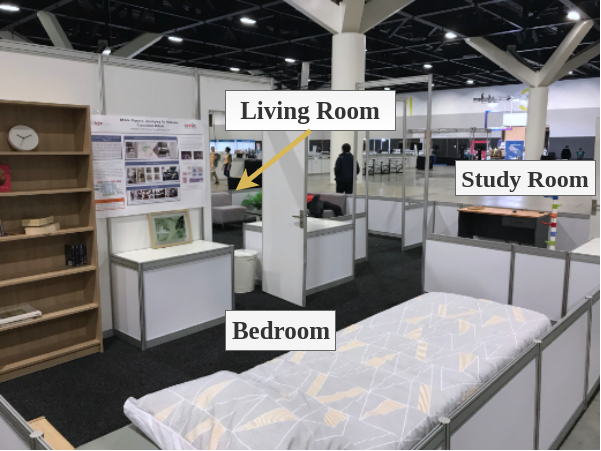}
    \caption{Mock apartment with the Bedroom and Study Room visible.}
    \label{fig:arena_photo2}
\end{figure}


\subsection{Storing Groceries}
In this task, a kitchen is set up with a table and a multi-shelf pantry cupboard, each of which contains common grocery items. After navigating to the kitchen table, the robot must pick up objects from the table and place them onto the pantry cupboard. A constraint, however, is that objects must be placed next to similar items. For example, the robot can place a box of orange juice next to a box of cereal since both are breakfast items. Another challenge of this task is its aggressive time limit of five minutes. As such, the storing groceries task requires fast perception and manipulation for general picking and placing of objects, as well as an understanding of human shelving preferences for determining where in the pantry cupboard to place objects.


\subsection{Take out the Garbage}
The goal of this task is for the robot to enter inside the arena when the entrance door is opened, navigate to two different trash cans located at known positions and transport the trash bags located therein to a designated deposit location.  For bonus points, the trash cans can start with lids on top that need to be removed.  As one of the most straightforward tasks in the competition, this task requires mainly navigation, manipulation of at least the trash bags and optionally the lids, and potentially rudimentary perception for fine-tuning of navigation and manipulation.  While there is an option to ask for help removing the lid, the HRI aspects of this task are minimal.  A particular challenge is that both bags must be deposited within five minutes, placing a premium on efficiency.  

\begin{figure}[t]
    \centering
    \includegraphics[width=0.5\columnwidth]{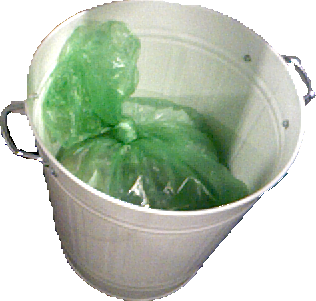}
    \caption{Trash can with the trash bag used at the arena.}
    \label{fig:trash_can}
\end{figure}


\subsection{Serving Drinks}
In the Serving Drinks task, the robot is to enter into a party and act as a waiter to deliver drinks to people at their request. The drinks are placed on a predefined bar located near the living room. There are a number of possible drinks, but only a subset of them are available for serving.
\begin{figure}[htb]
    \centering
    \includegraphics[width=0.8\columnwidth]{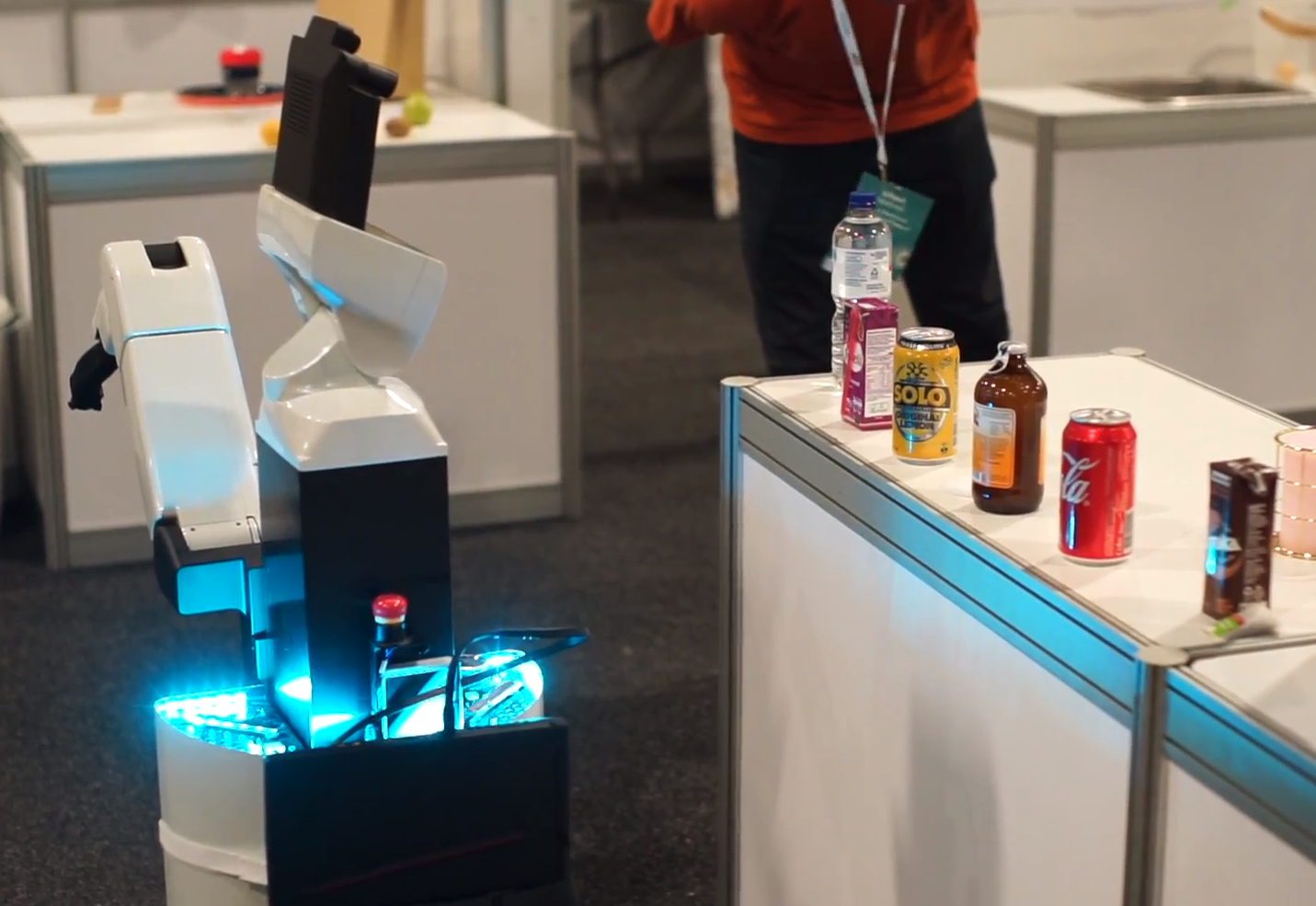}
    \caption{The HSR identifies available drinks at the bar.}
    \label{fig:drinks}
\end{figure}
There are several guests in the living room, three of which are sitting and the rest standing, two of whom already have drinks in their hands. The robot is to identify the guests that are in need of a drink, take their order, and deliver it. Guests are aware of the menu that is available, and, for bonus points, the robot can preemptively tell the patron that their order is not available. When the robot leaves the living room to retrieve a drink item for a guest, the guests are shuffled around to new positions. The task is aimed at requiring the robot to implement high-level HRI and combine it with object recognition and manipulation.

\subsection{Serve the Breakfast}
In Serve the Breakfast, the goal is for the robot to pour cereal into a bowl without spilling it. The robot begins outside of the kitchen, and must navigate to the kitchen when the door opens. Once inside, the robot must set the table or any flat surface. This requires the robot to navigate to the bowl, pick it up, and then place it gently on the table or a counter. For bonus points, the robot can place a spoon next to the bowl. Once the table is set, the robot must navigate to the cereal, pick it up, and return to the bowl to pour a bowl of cereal. For additional bonus points, the robot may pour milk into the bowl without spilling. 

\subsection{Carry My Luggage}
In this task, the robot helps the operator carry a bag to a car parked at an unknown location outside of the arena. At the start of the task, an operator points to a paper bag with handles. The robot can pick up the bag from the floor for extra points, or request a hand-over. The only way for the robot to reach the destination is by following the operator. The main challenge of this task is for the robot to keep track of the operator while its path is obstructed by other people, small objects, and barriers. 


\subsection{Restaurant}
Designed to capture the complexity of busy restaurants, the Restaurant task has the robot act as a waiter in a gastropub-style restaurant, ideally one serving real customers. The robot must detect the bar where food and drinks are requested by waiters and prepared by the kitchen. Customers can call for waiters, usually by waving or shouting, and the robot must move to these customers and take their orders. These orders should be propagated to the kitchen, after which, the robot should deliver the prepared meals from the bar back to the customers' tables. Many challenges of this task come from the fact that the restaurant is situated in the real world and not known beforehand. Thus, navigation must be done in a very crowded and unmapped environment. Furthermore, gesture recognition, perception, and manipulation should be developed enough to be able to detect and deliver orders.

\subsection{Clean the Table}
After a family dinner, the robot is tasked with clearing a dining table. This involves picking up plates and small spoons from a table and placing them in a dishwasher. Extra credit is given for opening the dishwasher and placing a detergent pod in the machine's soap slot. A challenge is the manipulation of small and uniquely shaped objects: cutlery and plates, respectively.

\section{Component AI Technologies}
This section describes the component technologies we developed across multiple tasks for our robot architecture, semantic perception, and object manipulation on top of the HSR software stack~\cite{Yamamoto2019}.

To the extent possible, we built our approach in a manner consistent with our ongoing Building-Wide Intelligence project~\cite{bwibots}.  While using a different hardware platform, many of the objectives and capabilities are the same.  Indeed we have previously designed an underlying architecture that is common to the two platforms~\cite{jiang2018laair}.

\subsection{Robot Architecture}

\begin{figure*}
    \centering
    \includegraphics[width=\textwidth]{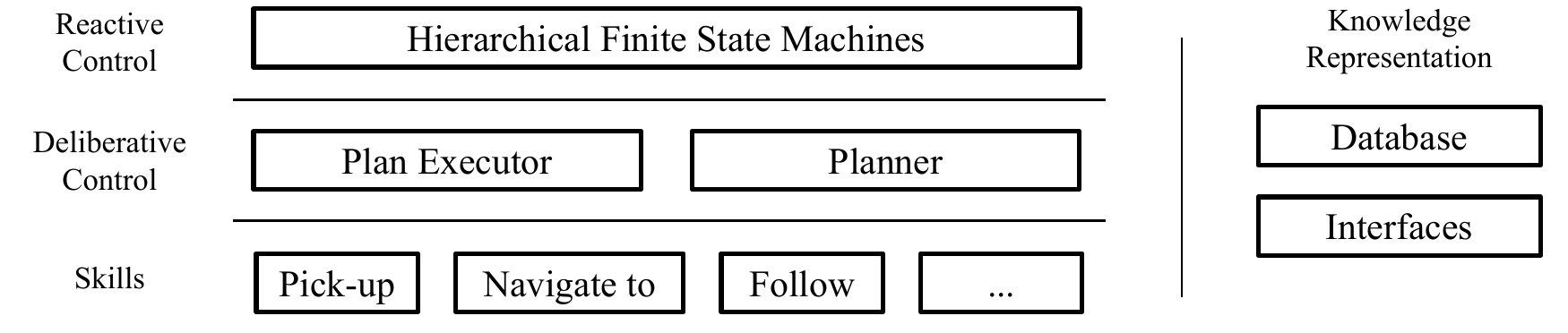}
    \caption{Implementation of our robot architecture on HSR.}
    \label{fig:hsr_system_diagram}
\end{figure*}

Our architecture is designed for service robots to handle dynamic interactions with humans in complex environments. The three-layer architecture, as shown in Figure~\ref{fig:hsr_system_diagram}, outlines integration of the robot's skill components, such as perception and manipulation, with high-level reactive and deliberative controls. The top layer sequences and executes skills, and is reactive during execution to respond to changes. A central knowledge base facilitates knowledge sharing from all the components. The deliberative control layer uses the knowledge base to reason about the environment, and can be invoked to plan for tasks that cannot be statically decomposed. Details on implementation of these layers can be found in our recent paper~\cite{jiang2018laair}.

\subsection{Perception}

We employ a semantic perception module whose purpose is to process raw video and depth data from the robot's sensors and extract information that can be processed by the manipulation, navigation, and knowledge reasoning modules. The main output representations are a query-able point cloud of objects in the environment and a partial 3D map of the world.

The main input to our semantic perception module is RGBD camera data. Compressed RGB and depth images from the robot are streamed to an offboard computer that runs the perceptual system. This image data is then consumed by finding objects via the YOLO object detection network \cite{redmon2016yolo9000}, and constructing a point cloud.

Next, semantic information about the world is synthesized in two main ways: a partial 3D environmental map and object cloud. For the former, regions of the point cloud corresponding to detected objects are fused together over time in a probabilistic Octree representation based on Octomap \cite{hornung13auro}, which allows for the realtime construction of a partial 3D map of the world. For the latter, point estimates of the locations of objects are stored in a KD-Tree and wrapped with an efficient querying interface that integrates with our knowledge representation system.

The synthesized semantic information is then made available to plugins in an event-based model, where a plugin can request access to semantic information that it wants to operate on. Plugins used include custom RANSAC edge detectors used to detect surfaces, and bounding box fitting on the 3D map for use in manipulation. 

A significant limitation is the partial nature of the 3D environmental map. Only a partial map is constructed due to the realtime processing constraint; namely, full views of the world cannot be stitched together at framerate using the Octomap technology. Alternatively, GPU-based techniques for combining full point clouds could potentially overcome this limitation, and thus provides a direction for future development. Benefits of having full 3D environmental maps include the ability to directly localize objects with respect to the robot.


\subsection{Manipulation}

The purpose of our manipulation system is to enable the pick up and put down of diverse objects of different shapes and sizes. Our manipulation stack consists of three main components which we describe below: grasp pose generation, parallel motion planning, and closed-loop correction.

Grasp poses refer to poses of the robot's gripper that allow it to pick up an object. First, our semantic perception system provides 3D bounding boxes for objects worth manipulating. Based on these bounding boxes, potential grasp poses are computed that place the gripper on the top of the object as well as on all sides, with multiple possible rotations of the wrist. Of these poses, invalid configurations are filtered out by projecting the gripper onto the object and seeing if there is a collision.

Once grasp poses are determined, motion plans need to be determined in order for the robot to achieve a desired grasp pose. In order to do this quickly, we employ a parallel motion planning architecture built on top of the Moveit framework \cite{coleman2014reducing}. Our motion planning architecture is comprised of primary and secondary nodes. The secondary nodes handle generating motion plans for each potential grasp pose, while the primary node coordinates and handles executing motion plans. Specifically, secondary nodes plan in parallel, and the first motion plan found is what is executed. The rationale behind this is that different grasp poses will require different yet unknown amounts of time for finding motion plans. Since motion planning takes a significant amount of time, reducing this bottleneck greatly speeds up the entire manipulation pipeline. Furthermore, the Moveit framework can sometimes crash when trying to find plans. In our setup, this problem is mitigated: If a secondary node dies from such a crash, then the other secondary nodes are still present, allowing the system to continue functioning.

Next, executing a motion plan precisely is usually not feasible. This is because, as the plan is executed, the software solely uses odometry to control its position and the resultant drift can cause errors in how much the robot thinks it has moved. To overcome this obstacle, we slightly modify desired grasp poses by having the gripper be some offset away from the object. This way, after a motion plan is generated and executed, the robot's gripper is close to the object, but there remains a small gap. We take advantage of this small gap by employing a proportional controller based on object detections from the robot's hand camera to correct for odometry drift. This practically means that the robot shifts slightly to align the gripper perfectly with the centroid of the object. The gap is then closed by moving in a straight line towards the object, leading to a successful grasp. 

\section{Task Approaches}

This section provides, for each task attempted, a summary of our approach, the challenges overcome, successes and limitations of the approach, and directions for future improvements.

\subsection{Storing Groceries}

Fast perception and manipulation are crucial in this time-constrained task, which has shaped our approach. First, the robot identifies and localizes the kitchen table by exploiting two known pieces of information: the location of the pantry cupboard and the height of the table. Namely, the robot navigates to the pantry cupboard, and then scans around for an edge that is exactly the height of the table. Next, as objects have been passively perceived throughout the previous step, our semantic perception system is queried for all objects that are on the table. Of these, a random object is chosen and pushed through our manipulation system which causes the robot to pick up the object. Subsequently, the pantry cupboard is localized in a similar way as the table, and we query for all objects that are currently in the pantry. The final component of the task is deciding where to place the grasped object in our gripper. The simplest case is when the knowledge base knows a priori that two objects are part of the same category (e.g. sprite and ginger ale). Otherwise, we use a word2vec \cite{mikolov2013distributed} model fine-tuned on a custom corpus to decide the similarity between our grasped object and the objects in the pantry.

Our manipulation system is designed to work with a variety of object shapes and sizes. However, increased speed and reliability can come from exploiting the fact that most objects are quite small. Specifically, complex motion planning could be abandoned in favor of positioning the robot's gripper above the table and executing a simple motion downwards until the gripper hits the table. While not all objects can be picked up this way, many reliably can.

A video of a run\footnote{https://youtu.be/HAAaP7vTfmY} shows the robot successfully executing the above behavior. 


\begin{figure}[t]
    \centering
    \includegraphics[width=1.0\columnwidth]{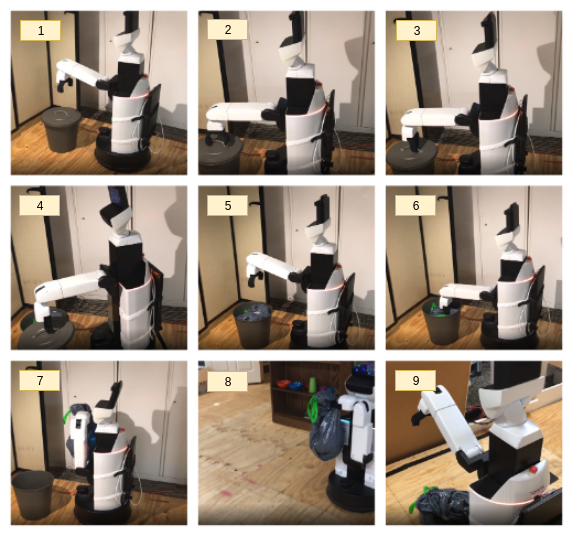}
    \caption{Sequence of actions for taking out a trash bag to the collection zone. Ordered from top left to bottom right.}
    \label{fig:trash_actions}
\end{figure}

\subsection{Take out the Garbage}
This task relies mainly on quick navigation and manipulation, with emphasis on speed and accuracy. A particular challenge is precise localization near the trash cans. As the locations of the trash cans are known beforehand, the HSR is able to navigate to a position in the arena where it is directly facing the trash can. Once facing the trash can, the HSR reaches out its arm and points its arm with the hand camera facing directly downward. From there, a 2D bounding box of the target is generated by YOLO object detection on the hand camera image. If the lid is on the can, it will be detected and become the target for the HSR to grab. Otherwise, the trash bag with the trash can will be detected and used as the grab target.

The HSR uses the position of the generated 2D bounding box to align its hand with the target. A proportional controller is used to publish a velocity command to the robot base based on the distance between the center of the hand camera image and the center of the bounding box. Once this distance is within a certain tolerance, the hand is directly above the target and the velocity command is set to zero. With the height of the trash can also known ahead of time, the HSR can then lower its arm straight down until it is at the height of the lid handle. If the lid is already removed, the HSR will instead lower its hand down into the trash can to grab the bag.

This strategy for grabbing the lid and bag worked consistently on multiple trials, so the primary concern of this task was completing it within the five minute time limit. Manipulation was made as efficient as possible by removing unnecessary movements and pauses in motion. Navigation speed was improved by setting waypoints in between destinations. These waypoints were strategically placed in open areas on the map, so the robot would not have to spend extra time navigating around known obstacles in the arena. The main drawback of this navigation strategy was that the robot would come to a complete stop at a waypoint before moving on to the next one, thus taking more time to reach the destination. Examples where this strategy worked are, in some runs, we noticed the robot reaching a state near doors where it gets to a pose from which it cannot plan its way through a door. To mitigate this issue, we place waypoints before and after the door frame, following which the robot is successfully able to navigate through open door frames. This strategy also ensures the robot always plans through the door in a straight path, avoiding collisions with the door frames and task interruptions due to bumper collision triggers. 

Figure \ref{fig:trash_actions} shows the sequence of states for the robot to pickup and deposit trash. In Step 1 , the robot navigates to the trash can and places its hand facing downward. In Step 2 and 3, the robot performs closed loop control using proportional controller to reduce the 2D translational error between hand camera plane and trash can lid plane. The robot then picks up the trash can lid, followed by the trash bag in steps 4,5,6 and 7. In step 8, the robot navigates to the collection zone, navigating through the arena avoiding obstacles along the way. Finally in step 9, the robot deposits the trash bag. We noticed that in step 8, the trash bag sometimes blocks the HSR's LIDAR and hence the navigation stack is unable to charter a plan to the goal as a blocked LIDAR is equivalent to a static obstacle infront of the robot. To solve this problem, we raise the height at which the HSR holds the trash bag while transporting it. In Step 9, we noticed the trash bag getting stuck on the gripper sometimes. To avoid this, we perform a bidirectional roll motion on the wrist, which helped in destabilizing the bag from the gripper and placing it successfully on the ground.

A video\footnote{https://youtu.be/Z8G2p7bkx3k} shows the above behavior including a pick up of one of the trash lids for extra points. 

\subsection{Serving Drinks}

This task presents perception and manipulation challenges alongside HRI. First, the robot navigates to the bar to check which drinks are available. Once this is done, we utilize OpenPose \cite{cao2018openpose} to detect and then navigate to people in the living room that require beverages. The closest person is asked for their name and drink order through Google Cloud's speech-to-text service. The speech recognition often misinterpreted the order or the name and to catch when a robot misheard their order, the person's speech was checked against a dictionary of rhyming words (e.g. to correct Santa to Fanta). However, this method has potential limitations should two drinks off the menu have similar names and improved recognition would facilitate human-robot interaction in this task. The robot proceeds to query our semantic perception system for the requested drink and then attempts to grab it. Ideally, the drink should be delivered to the same person that requested it, but facial recognition proved to be highly unreliable and therefore there was failure in delivering the drink back to the same person.

A video\footnote{https://youtu.be/ZBA3b2RTvfE} shows the robot's execution of the above behavior. 

\subsection{Serve the Breakfast}

This task challenges perception and manipulation capabilities. Using our semantic perception and manipulation systems, the HSR is able to recognize and pick up the bowl and cereal box located in the pantry cupboard. The bowl is first picked up and placed gently on top of the table. Using the known location of the table, a position near the close edge of the table is easily selected for object put down. The joint limits of the HSR compel a clever pouring configuration which required the hand to be upside down relative to the world vertical axis when picking up the cereal. To accomplish this, grasp poses from our manipulation system are sorted to remove those not fulfilling the aforementioned hand configuration requirement. Since only enough cereal to fill the bowl needs to be poured, the whole box cannot be emptied, which requires detecting how much cereal has been poured. One challenge is that the wrist force/torque sensor is quite noisy, which renders measuring the weight of cereal poured implausible. Adding to this challenge are the morphology of the HSR and the task time limit; the morphology forces the HSR to assume a side view to visually detect the amount of cereal poured. Performing this body rotation and computation adds significant time delay. Accurately measuring the amount of cereal poured is an essential development for meeting the demands of this task in the future.

\subsection{Carry My Luggage}
The key challenge of this task is to follow the operator in a crowded environment. We implement person tracking with a combination of leg tracking, OpenPose, clothes color matching, and waving detection in a tight loop. Before following, the robot takes a photo of the person and sets the target legs. Since the LIDAR has a wider angle than head camera, the leg tracker results are preferred. When the target's legs are no longer visible, the head turns around to reidentify the target. For each person detected by OpenPose, a similarity score is computed using the color histogram of the person's body region and the original photo of the target. If the highest similarity score is above a threshold, the person is reset as the target; otherwise, the robot asks the target to wave, and looks for the closest waving person. We use the behavior tree framework~\cite{colledanchise2017behavior} to coordinate following and searching for target. Our solution is able to follow and regain the target with fair responsiveness in uncrowded spaces. To achieve robustness in uncontrolled environments, more accurate person identification will be required. We have plans to explore better sensor fusion techniques and special-purpose neural networks for person re-identification.

\subsection{Restaurant}
As the most dynamic task, Restaurant requires navigation, perception, and manipulation in an unseen and chaotic environment. For increased reliability, we bypass all manipulation in this task and focus our efforts on the navigational and human interaction challenges this task has to offer. First, the bar is detected by asking the bartender to raise his or her hands. Next, waiting customers are detected by employing a velocity-based hand waving gesture classifier; arms from OpenPose skeletons are identified, and the velocity of the wrist relative to the shoulder is checked over a few frames. This allows the robot to see which customers are waving. Once detected, the customer must be maplessly navigated towards. 

A challenge however is determining \emph{where} to move to. After all, moving to exactly the customer's location would be equivalent to running them over, which would lead to immediate disqualification. Instead, just like a normal waiter, the robot should move close to the customer, such as right beside their table. Though, since we are in a previously unseen area, the robot has no knowledge of where tables are or what areas would be appropriate to move to. To that end, the robot looks at its local obstacle map, and finds the "island" the the customer is on. By island we mean an occupied region surrounded by free space. Generally, the customer, the chairs they are sitting on, and their table, will be an island surrounded by free space that the robot can move to. The shortest path to the customer is planned, and the farthest point on that path which does not collide with the island is where the robot moves to.

\subsection{Clean the Table}
Core challenges in this task include the manipulation of small and uniquely shaped objects. We chose to bypass picking up of objects, so we have the robot first navigate to the dining table and ask for cutlery and dishes to be handed over. Afterwards, we navigate to the known location of the dishwasher. Now, objects must be placed precisely into the dishwasher. Failure to do so for even a single object leads to disqualification, and so localization cannot be solely relied on for the dishwasher's location. Instead, we use localization for an initial rough estimate of where the dishwasher is. To correct for any error, the robot looks at its local costmap (which fuses LIDAR measurements and a projected point cloud) for the corner of the dishwasher, which it aligns itself to. Now certain of being reliable aligned to the dishwasher, the robot places an object in its gripper into the dishwasher rack.

\section{Related Work}


Our participation in RoboCup@Home has initiated several research efforts in AI and HRI in previous years. We developed the robot architecture for general purpose service robots and performed a case study on the HSR~\cite{jiang2018laair}. We proposed a knowledge representation and planning approach to handle human requests that involve unknown objects~\cite{ICAPS19-Jiang}. Further, we discussed challenges and synergies in building a unified system for RoboCup@Home and our custom office robot platform, BWIBot~\cite{bwibots}. This paper focuses on our approaches to tasks in the new rulebook in 2019.

The variety of rich tasks in RoboCup@Home have led to diverse approaches and research by other teams in the league. For one, Contreras et. al. describe a different system that team Er@sers' fielded in the clean the table task \cite{contreras2018multimodal}, where they use an active object interaction system with multimodal feedback. Savage et. al. present the architecture used by team PUMAS in which a layered architecture is combined with semantic modules for executing competition tasks \cite{savage2019semantic}. Team Northeastern has described their system for mobile manipulation, with a focus on its deployment in the Storing Groceries task~\cite{Kelestemur:2019}.

\section{Summary and Conclusion}

Overall, the current RoboCup@Home rulebook defines an inspirational stretch goal for modern human-interactive service robots.  This paper describes the UT Austin Villa 2019 approach to the full variety of challenges presented - both to the component AI and HRI technologies and to their integration.   While important progress has been made, as we document fully in the paper, there remains much room for improvement and we look forward to continuing our research and development towards the 2020 competition in Bordeaux!

\bibliographystyle{aaai}
\bibliography{references}

\end{document}